\documentclass[pdflatex,sn-mathphys-num]{sn-jnl}

\usepackage{graphicx}%
\usepackage{multirow}%
\usepackage{amsmath,amssymb,amsfonts}%
\usepackage{amsthm}%
\usepackage{mathrsfs}%
\usepackage[title]{appendix}%
\usepackage{xcolor}%
\usepackage{textcomp}%
\usepackage{manyfoot}%
\usepackage{booktabs}%
\usepackage{algorithm}%
\usepackage{algorithmicx}%
\usepackage{algpseudocode}%
\usepackage{listings}%
\usepackage{natbib}
\usepackage{tikz}
\usetikzlibrary{graphs}
\setcitestyle{authoryear,open={(},close={)}} 

\usepackage{times,latexsym}
\usepackage{url}
\usepackage[T1]{fontenc}
\usepackage{graphicx}
\graphicspath{ {./figures/} }
\usepackage{hyperref}
\usepackage{titlesec}
\usepackage{mathtools}
\DeclarePairedDelimiter\ceil{\lceil}{\rceil}

\usepackage{xstring}

\usepackage{color}
\usepackage{comment}
\usepackage{amsmath}

\DeclareMathOperator*{\argmin}{arg\,min}
\usepackage{makecell}

\theoremstyle{thmstyleone}%
\usepackage{color}
\definecolor{codegreen}{rgb}{0,0.6,0}
\definecolor{codegray}{rgb}{0.5,0.5,0.5}
\definecolor{codepurple}{rgb}{0.58,0,0.82}

\lstdefinestyle{python_code}{
    commentstyle=\color{codegreen},
    keywordstyle=\color{magenta},
    numberstyle=\tiny\color{codegray},
    stringstyle=\color{codepurple},
    basicstyle=\footnotesize,
    breakatwhitespace=false,
    breaklines=true,
    captionpos=b,
    keepspaces=true,
    numbers=left,
    numbersep=5pt,
    showspaces=false,
    showstringspaces=false,
    showtabs=false,
    tabsize=2
}
\theoremstyle{thmstyletwo}%

\theoremstyle{thmstylethree}%

\newcommand{\corpus}{Turkronicles}

\newcommand{\rgnat}{RGNAT}
\newcommand{\secref}[1]

\begin{document}

\title[Article Title]{\corpus: Diachronic Resources for the Fast Evolving Turkish Language}


\author[1]{\fnm{Togay} \sur{Yazar}}\email{u.yazar@etu.edu.tr}

\author[2]{\fnm{Mucahid} \sur{Kutlu}}\email{mucahidkutlu@qu.edu.qa}

\author[3]{\fnm{İsa Kerem} \sur{Bayırlı}}\email{isakerem@gmail.com}
\affil[1]{\orgdiv{Department of Computer Engineering}, \orgname{TOBB University of Economics and Technology}, \orgaddress{\city{Ankara}, \country{Turkey}}}

\affil[2]{\orgdiv{Department of Computer Science and Engineering}, \orgname{Qatar University}, \orgaddress{\city{Doha}, \country{Qatar}}}

\affil[3]{\orgname{TOBB University of Economics and Technology}, \orgaddress{\city{Ankara}, \country{Turkey}}}


\abstract{Over the past century, the Turkish language has undergone substantial changes, primarily driven by governmental interventions. In this work, our goal is to investigate the evolution of the Turkish language since the establishment of Türkiye in 1923. Thus, we first introduce \corpus{} which is a diachronic corpus for Turkish derived from the Official Gazette of Türkiye. \corpus{} contains 45,375 documents, detailing governmental actions, making it a pivotal resource for analyzing the linguistic evolution influenced by the state policies. In addition, we expand an existing diachronic Turkish corpus which consists of the records of the Grand National Assembly of Türkiye by covering additional years. Next, combining these two  diachronic corpora, we seek answers for 
two main research questions:  How have the Turkish vocabulary and the writing conventions changed since the 1920s? 
Our analysis reveals that the vocabularies of two different time periods
diverge more as the time between them increases, and newly coined Turkish words take the place of their old counterparts. We also observe changes in writing conventions. In particular, the use of circumflex noticeably decreases and words ending with the letters "-b" and "-d" are successively replaced with "-p" and "-t" letters, respectively. 
Overall, this study quantitatively highlights the dramatic changes in Turkish  from various aspects of the language in a diachronic perspective.}




\maketitle

\section{Introduction}\label{sec1}

Languages undergo perpetual transformations over time. This evolution can be attributed both to natural factors such as semantic bleach and simplification and to cultural factors such as technological advancements and social developments.  These changes might reduce the utilization of language models for historical texts due to differences in the meaning or spelling of words in their training data. Therefore,  it is important to understand the historical evolution of languages. 

Turkish language had a noticeably different path in terms of how it has evolved in the last century compared to other languages. In particular, after the establishment of the Republic of Türkiye\footnote{Previously known as Turkey, of which the official name has been changed to Türkiye in 2023} in 1923, cultural and technological modernization was the most urgent agenda of the reform program adopted by the new government. The Turkish language underwent two radical changes in the context of this new campaign towards modernization: one concerning its alphabetic system and one its lexical repertoire. In 1928, the Perso-Arabic script\footnote{\url{https://en.wikipedia.org/wiki/Ottoman_Turkish_alphabet}} in which Turkish had been written during the period of the Ottoman Empire was given up in favor of a version of the Latin script consisting of 29 letters. 
The second major change in the Turkish language was an attempt to "simplify" and "purify" the Turkish language by replacing words of Persian and Arabic origin, which were numerous during the Ottoman period, with words of Turkish origin (i.e., with words that are either historically Turkish or derivable by the rules of Turkish morpho-phonology). This process was concomitant with the establishment of the Turkish Language Association, widely known as TDK\footnote{\url{http://tdk.gov.tr}}, in 1932 and can be understood to be part of the attempts at the crystallization of a new national identity.



In this work, we investigate how the Turkish language has changed since 1920s. In order to conduct our study, we first developed   a diachronic corpus for Turkish. Specifically, we crawled issues of the Official Gazette of Türkiye (OGT) and the records of Grand National Assembly of Türkiye between 1920 and 2022. 
Given that both  resources contain documents about governmental actions such as laws, regulations and discussions surrounding them,  we think that our diachronic corpus is an important resource for analyzing the evolution of the Turkish language and the government's role in this transformation. 
Our corpus contains 45,375 documents, 842M words and 211K unique words. Using our corpus, we seek answers to the following research questions. 

\textbf{RQ-1: How has Turkish vocabulary changed since 1920?} We 
divide our dataset into ten-year periods and compare the words used in each time period \textcolor{black}{using different methodologies}. We find that the words in two time periods diverge more as the time difference increases. While the frequency of the newly coined Turkish words increases over time, the frequency of their counterpart words with Arabic of Persian origins decreases. Around  75\% of the words existing in 1920s have not been used between 2010 and 2019.

\textbf{RQ-2: How has the writing conventions changed since 1920s?} We observe the use of circumflex noticeably has decreased noticeably  compared to the 1920s and 1930s. In addition, the last letter of the words changed over time based on Turkish phonology. In particular, we found that the use of words that end with ``-b'' (e.g., ``kitab", which means book) decreases over time compared to its versions in which the last letter is ``-p'' (i.e., ``kitap"). However, we observe a different pattern for words ending with ``-d/t'' letters: The percentage of words ending with  ``-d'' compared to the corresponding words with ``-t'', e.g., Ahmed vs. Ahmet in 2010s, is similar to their percentage in 1920s, although it has a decreasing trend since 1990s.

The main contributions of our work are as follows. 
\begin{itemize}
    \item We create the largest \textcolor{black}{Turkish diachronic resources comprised of a diachronic corpus of formal Turkish documents, different kinds of diachronic word embeddings, a digitalized modern-old Turkish counterparts dictionary, and a python library that enables diachronic analysis.}   
    \item We explore the language change in Turkish since the 1920s using our corpus. 
    \item We share our code and data to enable further research studies\footnote{URL is hidden due to the double-blind review process}.
\end{itemize}

The rest of the paper is organized as follows. We first provide background information about Turkish for non-Turkish speakers in Section \ref{sec_background}. Next, we discuss the related work in Section \ref{sec_related_work}. In Section \ref{sec_dataset}, we explain how we constructed our \textcolor{black}{diachronic resources}. We present our analysis and discuss our findings in Section \ref{sec_data_analysis}.  In Section \ref{sec_limitations} we discuss the limitations of our work and conclude in Section \ref{sec_conclusion}.

\section{Background} \label{sec_background}
Turkish belongs to the southwestern/Oghuz branch of the Turkic language family which includes languages such as Uigur, Uzbek, Kazakh, and Kyrgyz \citep{Johanson1998Turkic}. Its characteristic phonological feature is the assimilatory process of vowel harmony whereby a vowel shows agreement with the preceding vowel in terms of frontness and, to a more limited extent, roundness \citep{lees1961phonology}. Moreover, Turkish obeys various phonotactic constraints such as the absence of adjacent vowels inside words (with the exception of the “loan” words) as well as the ban on word-final voiced stop consonants such as [b], [d] and [g] (again, with the exception of some mono-morphemic words such as \textit{ad}, meaning ``name"). At the morphological level, Turkish is known as an agglutinative language in which inflectional suffixes attach to a nominal or verbal stem one by one, creating a structure similar to beads on a string. Syntactically, the default word order in Turkish is Subject-Object-Verb (SOV); however, other word order permutations are also acceptable under various prosodic and information-structural conditions, especially in spoken registers (see \cite{lewis2000turkish}, \cite{underhill1976turkish};  \cite{Kornfilt1997Turkish}; \cite{goksel2004turkish} for extensive overviews of Turkish grammar).

\section{Related Work}\label{sec_related_work}
In this section, we discuss the studies in the literature from two different perspectives parallel to the contributions of this study. 

\subsection{Turkish Corpora}
The NLP resources for Turkish are highly limited compared to English. However, the digitization of printed materials enabled the development of various Turkish corpora.
METU Corpus \citep{say2002development}, and Turkish National Corpus (TNC) \citep{aksan2012construction} are general-purpose, genre-balanced, Turkish corpora. Both accommodate text resources from different genres, and the period of the text files is post-1990. However, they differ in the size of word count and annotation style. The former has 2 million tokens and XCES annotations, while the latter has 50 million words and provides part-of-speech tag annotations. 

There are also larger Turkish corpora available such as BOUN Corpus \citep{sak2008turkish}, with 500 million tokens derived from web. Moreover, METU-Sabanc{\i} Tree-bank \citep{say2002development} and IMST Turkish Dependency Tree-bank \citep{sulubacak2016imst} provide richer syntactic annotations such as morphological features, and dependency relations. However, none of these datasets clearly reflect language change, and hence, do not allow diachronic analysis. 

To the best of our knowledge, there exists only one diachronic corpus for Turkish which consists of documents of parliamentary sessions between 1920-2015 \citep{gungor2018corpus}, named Corpus of Turkish Grand National Assembly. 
Since the reports are exact transcribed versions of the speeches made by the deputies, the corpus can reflect the historical evolution of the modern Turkish language. 
In our work, we developed a more comprehensive corpus by extending the temporal scope of parliamentary records up to 2022 and crawling issues of the Official Gazette of Türkiye published between 1922 and 2022.  
We believe that the resulting corpus provides better insights into the linguistic dynamics of Turkish and the political discourse of Türkiye throughout its history.

\subsection{Diachronic Analysis} 

There are several  studies that consider various aspects of the language change in a diachronic perspective in the literature. \citet{michel2011quantitative} and \citet{lieberman2007quantifying} focus on the evolutionary dynamics of English. They explore the grammatical changes through the history and attempt to reveal long-term patterns in linguistic change and the effects of cultural shifts. Their studies are mainly built on quantitative analysis of the  frequency of the words across different time periods. As a different methodological approach, \citet{pechenick2015characterizing} utilize information theory methods, such as Jensen-Shannon Divergence, to examine the evolution of the English language by exploiting the Google Books dataset introduced by \citet{michel2011quantitative}. 

Many researchers use word embeddings to 
study semantic change throughout the years 
and reveal various facets and features of semantic change \citep{hamilton2016diachronic,szymanski2017temporal}. In addition, the meaning-bearing nature of the word embeddings allows the evaluation of the validity of linguistic hypotheses about semantic change, such as the Law of Parallel Change and the Law of Differentiation  \citep{xu2015computational}.

Although there exist several corpus-based diachronic analyses for English in the literature, the studies for Turkish are limited. These studies diachronically examine various aspects of the language. \citet{salan2022ccaugdacs} and \cite{vahit49turkccede} provide  extensive reviews on sound change in word-initial vowels of Turkish words and find instances of the phenomena by inspecting the dictionaries of different languages. \citet{sultanzade2012syntactic} qualitatively and quantitatively examine the valency change of a list of verbs in the Book of Dedekorkut by comparing them with modern Turkish correspondents. \citet{aksan1965turk} and \citet{bahattin2003anlam} study semantic change in Turkish and primarily investigate the development of individual words to explain the mechanisms causing the semantic shift.  To our knowledge, the study of \citet{gungor2018corpus} is the only work that analyzes the Turkish language in a corpus-driven diachronic way. They investigate the language change in their corpus by examining the frequency changes of near-synonym words and topic distributions using Latent Dirichlet Allocation. 
In our work, we use different computational approaches to observe the change in the lexicon and writing conventions using a larger corpus.

\section{\corpus{}}\label{sec_dataset}



\begin{table*}[!htb]
\small
\begin{tabular}{|p{0.08\linewidth}|p{0.86\linewidth}|}
\hline
\textbf{Date} & \textbf{Original}  \\
\hline
February 7, 1921 &  (1) Hâkimiyet bilâ-kayd ü \c{s}art milletindir. \.{I}dare usulü, halk{\i}n mukadderat{\i}n{\i} bizzat ve bilfiil idare etmesi esas{\i}na müstenittir. (2) Türkiye Devleti, Büyük Millet Meclisi taraf{\i}ndan idare olunur ve hükûmeti “Türkiye Büyük Millet Meclisi Hükûmeti” unvan{\i}n{\i} ta\c{s}{\i}r.  \\

&\textit{(1) Sovereignty unconditionally belongs to the people. The administration is based on the principle that the people themselves directly and actively manage their own destiny. (2) The State of Turkey is governed by the Grand National Assembly and its government bears the title of "Government of the Grand National Assembly of Turkey"}. \\ \hline
 October 4, 2020 &   (1) Bu yönetmeli\u{g}in amac{\i}, TOBB Ekonomi ve Teknoloji Üniversitesi Laboratuvar Okullar{\i}ndaki e\u{g}itim-ö\u{g}retim, yönetim, kay{\i}t-kabul, devam-devams{\i}zl{\i}k, nakil ile ö\u{g}renci ba\c{s}ar{\i}s{\i}n{\i}n tespiti ve i\c{s}leyi\c{s}e yönelik usul ve esaslar{\i} düzenlemektir.
 (2) Laboratuvar okullar{\i} ile Üniversitenin ö\u{g}renci, ö\u{g}retim eleman{\i} ve ö\u{g}retmenleri birbirlerinin dersleri ile kültür, sanat, spor ve sosyal faaliyetlerine kat{\i}larak mü\c{s}terek etkinlikler gerçekle\c{s}tirirler.   \\

 &\textit{(1) The purpose of this code is to regulate the procedures and principles regarding education, management, registration-acceptance, attendance-absence, transfer, and determination of student success, as well as the operation of TOBB University of Economics and Technology Laboratory Schools. (2) Students, faculty members, and teachers from the Laboratory School and the university participate in each other's courses, as well as cultural, artistic, sporting, and social activities.}\\
\hline
\end{tabular}
\caption{Example sentences occurred from the official gazette of Türkiye from 1920 and 2022. The English translations are given in \textit{italics}.}
\label{tab_sample}
\end{table*}

As the Turkish state takes an active role in changing the Turkish language, the official statements made by the government might be one of the best ways to observe how the Turkish language is affected by the state. Therefore, we create \corpus{} which is the first diachronic dataset using the official gazette of Türkiye (OGT), named ``Resmi Gazete" and the official records of the Grand National Assembly of Türkiye (RGNAT). In this section, we first provide brief information about OGT and RGNAT. Next, we explain how we crawled the data  and prepared it for analysis. 

\subsection{Official Gazette of Türkiye} \label{sec_ogt}

OGT was founded on 7 October 1920 to inform about governmental actions and other topics such as statesmen's opinions about various issues. Its publication frequency has varied, from weekly to more sporadic schedules. 
Today, it is published every day except the holidays and Sundays based on the regulations implemented in 2009. 

The content of OGT is a reflection of the administration process of the Türkiye. The issues of OGT mostly contain  state-related news such as:
\begin{itemize}
    \itemsep0.5em
    \item Laws,
    \item Decisions of the Turkish Grand National Assembly and its internal regulations
    \item International treaties,
    \item Procedures about the dismissal, election, appointment, substitution, or resignation of the authorities such as vice president and high judicial members, ministers,
    \item Decisions regarding assignments, dismissals, and terminations of duty made by the president, 
    \item Interior administrative decisions such as administrative jurisdiction changes and decisions regarding the establishment of municipalities.
\end{itemize}

As official documents often take place in the gazette, yielding a formal language with almost no typos and grammar errors. The documents might also include non-Turkish texts due to the obligation to publish international agreements. 
In addition, the first 1053 
issues are originally written with the Ottoman Turkish alphabet. However, with the reform of the Turkish alphabet in 1928, OGT began using Latin letters\footnote{The issues with the Ottoman Turkish alphabet have been translated into modern Turkish in 2020 to honor the $100^{th}$ anniversary of OGT.}. Furthermore, they might contain non-sentence structures such as charts, tables, and others. \textbf{Table \ref{tab_sample}} provides example sentences extracted from OGT. 

\subsection{Extended Corpus of Grand National Assembly of Türkiye} \label{sec_rgnat}
As mentioned above, we have extended the temporal coverage of \citet{gungor2018corpus}'s corpus from 1920-2015 to 1920-2022. 
RGNAT consist of texts about all activities that occur during a session of the Grand National Assembly of Türkiye, including any kind of speeches, inspections, voting, noise, debates, schedules, agendas, reports, letters, and suggestions, which have been systematically documented since 1920. 
Since the meeting schedule of the parliament is variable from year to year, the publication frequency of these documents is irregular compared to OGT. 

 RGNAT and OGT have considerable overlap in terms of structural elements, such as charts and tables, and the topics covered. However, 
 unlike OGT, 
 parliamentary records capture a range of language styles from formal to colloquial, depending on the speaker and context. 

It is worth mentioning that, as in OGT, documents that belong to 1920-1928 are written in the Ottoman-Turkish alphabet. However, translated versions of these documents are available in the official website.

\subsection{Crawling} \label{sec_crawling}

All the published documents of OGT and \rgnat{}  are available at 
\url{www.resmigazete.gov.tr} and \url{https://www.tbmm.gov.tr/Tutanaklar/TutanakMetinleri}, respectively. 
To collect the documents, we used  Scrapy\footnote{https://scrapy.org} which is an open-source web scraping Python library. 
Using Scrapy's HTML parsing engine, we store the content of the documents extracted from the related web pages and their metadata. 
The metadata contains the publisher, publication date, file name, download link, and volume info. Eventually, we crawled issues of OGT published between  7 February 1921 and 31 December 2022, yielding 31,999 issues. For \rgnat, we collected 13,376 documents  published between  23 April 1920 and 1 August 2022. 

 \subsection{Preprocessing} \label{sec_prepare}




Most of the downloaded files are in PDF format.
We convert these pdf files into plain text files using PyPDF\footnote{https://pypi.org/project/pypdf} tool to easily process the documents. The issues of OGT with issue numbers between 24,092 and 28,500 are shared as texts directly, instead of PDF files. For these documents,  we directly extract the text content of these issues from the associated web pages.

We manually investigated the text extracting performance of PyPDF from  the documents. We observed that the tool is generally successful but its performance is degraded in three cases: i) poorly scanned documents, ii) physically damaged documents, and iii) documents with non-sentence elements such as tables and charts. 

In order to eliminate the errors introduced by the extraction tool and prepare the documents for further analysis, we perform the following pre-processing steps.

i) We reduce multiple consecutive spaces  to a single space and convert different types of space characters (e.g., tab  and non-breaking space characters)  to a single  regular space.

ii) We remove characters that cause problems in tokenization and/or do not have a representation in UTF encoding such as \textbackslash{xa0} and \textbackslash{xad}. 

iii) We use NLTK tool for tokenization. 

iv) In our manual processes, we observe that the incorrect extraction from PDF documents yield very uncommon words. Thus,  in order to eliminate those noisy ones, we remove 
words that contain any non-alphabetic characters and words that appear less than a particular threshold value. However, we realized that the quality of pdf files and, thereby, the performance of PyPDF changed over the years. Therefore, instead of using a single threshold value for all documents, we divide the data into ten-year time periods and we set the threshold value to  $\ceil*{\frac{N}{10000000}}$ where $N$ is the number of tokens of the time period under consideration.   We filtered out the words below the frequency threshold. We observed that this type of filtering mechanism was effective in removing noisy words.

v) As Turkish is an agglutinative and morphologically rich language, analysis of surface-level words might be misleading.  In order to detect lemmas, we use a morphological parser proposed by \citet{ozturel2019syntactically}.  If the morphological parser cannot find a stem for a particular word, we apply the F5 method (i.e., using the first five letters as stems) which is shown as an effective stemming method by \citet{can2006first}.

\subsection{\textcolor{black}{TDK Dictionary}}
We extract the modern Turkish equivalents of old Ottoman words from the 'Türkçeden Osmanlıcaya Cep Kılavuzu' \citep{sozluk} (Pocket Guide from Turkish to Ottoman Turkish), published in 1935 by the Turkish Language Association (Türk Dil Kurumu). This book aids speakers of contemporary Turkish in understanding and translating into Ottoman Turkish. It lists a wide range of modern Turkish terms alongside their Ottoman equivalents, presented in a certain format. Each entry is organized such that different senses of polysemous words, synonyms, terms, abbreviations, and French equivalents are separated and indicated with specific markers, which are explained on the book's opening page. First, we converted the PDF version of the book into a plain text file through \textit{tesseract} OCR (Optical Character Recognition), and subsequently, we employed regular expressions to generate a JSON file containing the entries. There are a total of 8647 newly coined Turkish words and their old counterparts available.

\subsection{\textcolor{black}{Ngrams}}
We have extracted unigrams, bigrams, and trigrams from each file, along with their frequencies. Preprocessing steps were applied to the tokens, as in the construction of other resources. Both surface-level forms and lemma frequencies can be readily accessible. In addition, we've organized the n-grams into the time periods and marked them with timestamps. One can analyze the distribution of these ngrams to examine the change in the usage of individual words or phrases through time. Moreover, we provide the association strength of unigrams using the PPMI measure. This measure provides insight into the strength of association between unigrams, allowing for a deeper understanding of language usage patterns and relationships. 

These resources enable users to easily conduct queries pertaining to frequency changes over time, discover linguistic patterns, and test hypotheses about the Turkish language across different time periods.

\subsection{Lingan: A Python Library for Linguistic Analysis}
We developed a Python library to conduct diachronic analyses and facilitate the reproducibility of our experiments. 

\subsubsection{Components of Lingan}

Lingan is fundamentally based on three different layers of abstractions: \emph{Data}, \emph{Container}, and \emph{Operation}.
The \emph{Data} component is a representation of actual data that is used in linguistic analysis. Practical equivalents of this component include data types such as word embeddings, vocabulary, and ngrams. Members of this class contain both static and derived features of the relevant data. Generally, the responsibility of this component is to manage the data.

\begin{figure}[htbp]
\centering
\begin{tikzpicture}[sibling distance=15em,
  every node/.style = {shape=circle, draw, align=center, minimum size=3em, }]

  \node {$D_1$}
    child { node {$C_1$} }
    child { node {$D_2$}
      child { node {$C_3$} }
      child { node {$C_4$} } 
      };

\end{tikzpicture}
\vspace*{5mm}
\caption{An example of a hierarchical tree structure consisting of Corpus and DiachronicCorpus objects. Nodes with \emph{D} and \emph{C} represent \emph{DiachronicCorpus}, respectively. \emph{$D_2$} contains two \emph{Corpus} objects, \emph{$C_3$} and \emph{$C_4$}. $C_1$ and $D_2$ together compose $D_1$.}
\label{fig:tree_diagram}
\end{figure}
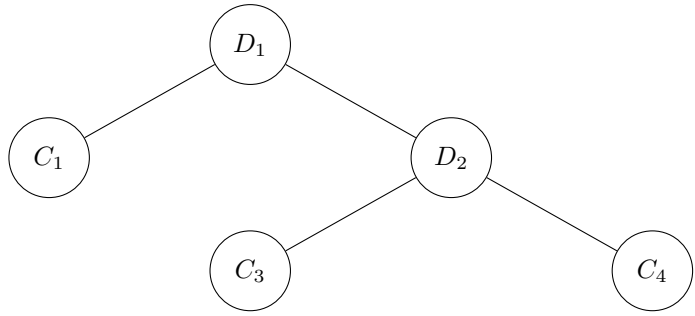

\emph{Container} represents collectively created text data. Technically speaking, the \emph{Container} component is a common interface for the hierarchical structure of the container classes used to wrap objects representing data,  meaning that data objects should be wrapped by a Corpus object to perform diachronic analysis. Within Lingan, there are two classes derived from this interface, namely \emph{DiachronicCorpus} and \emph{Corpus}. \emph{Corpus} class also includes attributes to interact with text files. For example, \emph{TokenProcessor}, which sequentially transforms tokens, and \emph{TextProcessor}, which is in charge of processing streamed raw text data, play an active role within this class. On the other hand, \emph{DiachronicCorpus} is a composite object that consists of many \emph{Container} objects, such as \emph{Corpus} and \emph{DiachronicCorpus}, marked with a timestamp of their time periods. \emph{Corpus} and \emph{DiachronicCorpus} together constitute a tree structure. An example of such a tree is depicted in Figure \ref{fig:tree_diagram}.

The Operation component is responsible for the algorithms performed on the hierarchical structure of a \emph{Container}. Additionally, the \emph{Data} objects should be generated by subclasses of \emph{Operation}. This is a design decision ensuring system-wide integrity and consistency. Also, operations are separated from the data structure, \emph{Container}, to enhance flexibility and maintainability, e.g., without modifying the composite structure, one can easily define new operations. However, the design of \emph{Operation} enforces the classes derived from \emph{Operation} to implement functions specific to the \emph{DiachronicCorpus} and \emph{Corpus} types. In Lingan, we have implemented the functionality used in the section \ref{sec_data_analysis}. The functions are readily available to the users of the library. We share a sample piece of code below to show the easy usage of the predefined operations. The task to be performed is to calculate the relative frequency of a specific word.

\begin{figure}[htbp]
    \centering
    \begin{lstlisting}[language=Python]
corpus_1930 = Corpus.load("path/to/corpus_1930")
corpus_1980 = Corpus.load("path/to/corpus_1980")

dia_corpus = DiachronicCorpus(corpora=[corpus_1930, corpus_1980])
frequency_series = dia_corpus.perform(Frequency(word = "belge", normalize=True))
print(frequency_series)
\end{lstlisting}
\caption{An example usage of Lingan. This code piece computes the relative frequency of \textit{belge} (document) across time periods through pre-defined function \textit{Frequency}.}
\label{fig:fig_code_main}
\end{figure}

The code assumes that \emph{Corpus} objects are serialized and saved locally. In the first two lines of the program, corpus of 1930–1939 and 1980–1989, the variables \emph{corpus\_1930} and \emph{corpus\_1980}. In line 4, a diachronic corpus is initialized by constructing the tree structure mentioned above. In line 5, a new \emph{Frequency} instance, which is the concrete object responsible for calculating the relative frequency of a given word, is created and passed as an argument to the \emph{perform} method of \emph{dia\_corpus}. The result is stored in the variable \emph{frequency\_series}.

\subsubsection{Available Operations}
We have implemented various data structures, containers, and operations that can be utilized in a diachronic analysis. Out-of-the-box functionality of the library is listed in Table \ref{tab:lingan_functions} with their names, data structures, on which the operations are performed, and short descriptions. In implementing these functions and other parts of the framework, libraries such as \emph{numpy} and \emph{scipy} were extensively used. Note also that the entire operations in the framework are not listed in \ref{tab:lingan_functions}; There are also different types of operations such as \emph{CreatePPMIMatrix, CreateSVDEmbeddings, CreateNgrams, CreateVocabulary} which create data models defined in the framework.

\begin{table}[htbp]
    \centering
    \begin{tabular}{p{0.40\linewidth} | p{0.15\linewidth} |p{0.40\linewidth}}
      \textbf{Operation}  & \textbf{Datastructure} & \textbf{Description} \\ \hline
       \emph{Exists(word, time\_range)} & \textit{Vocabulary} & Checks whether \textit{word} exists in a diachronic corpus within the time range \textit{time\_range}.  \\
       
       \emph{Frequency(word, time\_range)} & Vocabulary & Returns a time series where each member is the frequency of \textit{word} in each corpus within the \textit{time\_range} \\
       
       \emph{MergeVocabulary(time\_range)} & Vocabulary & Merges the vocabularies in the \textit{DiachronicCorpus} and returns the composed vocabulary \\ 
       
       \emph{FilterFrequency(threshold, time\_range)} & \textit{Vocabulary} & Filters the vocabulary of each corpus, i.e, removes the words whose frequency is below  \textit{threshold} \\
       
       \emph{VocabularySimilarity(time\_range)} & Vocabulary & Returns a matrix where each element results from Jaccard Index between the vocabularies of different time periods. \\
       
       \emph{VocabularyDistance(time\_range)} & Vocabulary & Returns a matrix where each element represents Jensen-Shannon divergence between different time periods\\
       
       \emph{MorphemFrequency(morpheme, time\_range)} & \textit{Vocabulary} & Returns the usage frequency of \textit{morpheme} within the \textit{time\_range}, e.g., the usage of \textit{â} across the time periods \\
       
       \emph{WordsWithMorpheme(morpheme, time\_range)} & \textit{Vocabulary} & Returns the words in each time periods that contains \textit{morpheme} \\
       
       \emph{WordsEndWith(suffix, time\_range)} & \textit{Vocabulary} & Returns the words in each time period that ends with \textit{suffix} \\
       
       \emph{WordsStartsWith(prefix, time\_range)} & \textit{Vocabulary} & Returns the words in each time period that starts with \textit{prefix} \\
       
       \emph{UniqueWordCount(time\_range)} & \textit{Vocabulary} & Returns a time series where each element is the total number of unique words in the vocabulary of each time period. \\
       
       \emph{CommonWords(time\_range)} & \textit{Vocabulary} & Returns a set of words that exist in every time period \\
       
       \emph{AverageWordLength(time\_range)} &  \textit{Vocabulary} & Returns an array of numbers where each element represents the average length of unique words in each time period \\
       
       \emph{NgramCount(time\_range)} & \textit{Ngrams} & Returns a time series data of the total number of ngrams in each time period. \\
       
        \emph{CoFrequency(u, v, time\_range)} & \textit{Embeddings} & Returns a time series of cooccurence frequency of the word \emph{u} and \emph{v} within the \textit{time\_range}. \\
        
       \emph{Collocations(word, k, time\_range)} & \textit{Embeddings} &Returns an array where each element is a set of words. These sets with size \textit{k} correspond to the words with the highest collocation value with the \textit{word} for each period \\
       
       \emph{Association(u, v, time\_range)} & Embeddings & Returns a time series array. Each element in this array corresponds to the collocation value between the words \textit{u} and \textit{v} in the respective time period. \\

       \emph{Similarity(u, v, time\_range)} & Embeddings & Returns a time series array where each element is the similarity between \textit{u} and \textit{v} in respective time periods \\

       \emph{AlignedMostSimilar(word, k, target\_period, base\_period)} & Embeddings & Returns \textit{k} most similar words to \textit{word} by aligning the embedding space of \textit{target\_period} to that of \textit{base\_period} \\

       \emph{SemanticChange(word, time\_range)} & Embeddings & Returns a time series array where each element is the distance from the vector of \textit{word} in the starting period. \\

       \emph{MostSimilar(word, k, time\_range)} & Embeddings & Returns an array where each element corresponds to a specific time period and the \textit{k}  closest words to \textit{word}  in that time period\\
       
       \hline
    \end{tabular}
    \caption{The list of currently available operations in the library}
    \label{tab:lingan_functions}
\end{table}

\subsubsection{Extensibility}

Due to the flexible nature of Lingan, our framework can be easily extended in terms of the fundamental components of the architecture. Users can define their custom data types and operations. Custom types should conform to the related interface. The proper way to achieve this is to extend relevant classes and interfaces. For clarity, we provide a showcase example where the task is to calculate the total number of sentences in a diachronic corpus. However, the data structure and the function for this task are not defined in Lingan. One should first define the data model:

\begin{figure}[htbp]
\begin{lstlisting}[language=Python,]
class Sentences(Data):
    def __init__(self, sentences: list[str], *args, **kwargs):
        super().__init__(*args, **kwargs)
        self.sentences = sentences
    
\end{lstlisting}
\caption{Defining a new \textit{Data} component to model sentences in the corpus.}
\end{figure}

The data model inherits from \emph{Data} interface to specify that \emph{Sentences} is a newly defined data structure in the framework. Next, the logic for counting sentences is implemented in a subclass of \emph{Operation}.

\begin{figure}[htbp]
\begin{lstlisting}[language=Python]
class NumberOfSentences(Operation): 
    def on_corpus(self, corpus: Corpus) -> int:
        return len(corpus.sentences)

    def on_diachronic_corpus(self, diachronic_corpus: DiachronicCorpus) -> int:
        total = 0
        for c in diachronic_corpus:
            total += c.perform(self)
        return total

\end{lstlisting}
\caption{Defining a new \emph{Operation} to calculate the total number of sentences on a diachronic corpus structure.}
\end{figure}

\textit{NumberOfSentences} contains two methods: \textit{on\_corpus} and \textit{on\_diachronic}. These are abstract methods from \textit{Operation} interface that every subclass should implement to interact with the composite \textit{DiachronicCorpus} structure. NumberOfSentences class traverses each element of the Corpus in a tree structure one by one, and defines its operation recursively according to the type of each element. Here in line 7, there is a method named \emph{perform} and bounded to \emph{Corpus} object. This is one part of the double dispatch mechanism of the framework: \emph{perform} method of currently processed \emph{CorpusContainer} object is invoked with an \emph{Operation} instance as its argument, and inside the \emph{perform} method CorpusContainer object \emph{c} calls one of \emph{on\_diachronic\_corpus} and \emph{on\_corpus} according to its type and passes itself as an argument, e.g., if the element is an instance of the Corpus class, then it invokes the \emph{on\_corpus} method of the operation.

\subsection{\textcolor{black}{Embeddings}} \label{sec_embeddings}
We provide three types of diachronic embeddings: PPMI (Positive Pointwise Mutual Information), SVD (Singular Value Decomposition) of PPMI, and CBOW (Continuous Bag of Words) embeddings. First, preprocessed text files are grouped into 10-year time periods according to their publication date. For each period, we count term-to-term cooccurrences to construct the PPMI matrix. The following formula is used to fill the elements of the matrix:

\begin{equation}
PPMI(u, v) = \max\left(\log \frac{p(u, v)}{p(u) \cdot p_{\alpha}(v)}, 0\right)
\end{equation}

where $p(u)$ and $p(u,v)$ are the marginal probability of word $u$ and the joint probability of words $u,v$ respectively. It is well known that PPMI is very sensitive to infrequent events. So, $p_{\alpha}(v)$, smoothed unigram distribution \citep{mikolov2013efficient} of word $u$, with $\alpha=0.75$ is used to alleviate such negative effects. Also, an unweighted context window with size 2 was employed to relate the target word to the context words. After that, SVD factorization of PPMI matrices has been taken. In the SVD approach, we calculated the vector representations of the words by $W=U\Sigma^{1/2}$  where $U$ is left singular vectors and $\Sigma$ is the singular values of the PPMI matrix. The size of the embeddings is 300. Note that, unlike the classical SVD implementation, we take the square root of $\Sigma$ which has been shown to improve the quality of the SVD embeddings \citep{levy2015improving}.

Finally, CBOW embeddings \citep{mikolov2013efficient} are created using \textit{gensim} \citep{vrehuuvrek2010software} library for each time period. We use context window size of 2, $alpha=0.75$, and embedding size = 300 for the CBOW algorithm as in SVD embeddings. Furthermore, CBOW-specific parameters such as the number of negative words and downsample rate are chosen to be 5 and 0,00001. 

The non-unique nature of SVD and the randomized processes involved in CBOW embeddings, prevent direct comparison of embeddings from different periods \citep{hamilton2016diachronic}. Word embeddings for the same word, trained at distinct times and with the same parameters, can still be different from each other. Specifically, two embedding spaces may be rotated, translated, or dilated via a transformation matrix $R$. Therefore, we align the embedding matrices $W_{t_1}$ to $W_{t_2}$ using the Orthogonal Procrustes Problem \citep{schonemann1966generalized}. The solution to the Orthogonal Procrustes problem tries to find a transformation matrix R that minimizes the Frobenius norm of the squared difference of two embedding matrices:

\begin{equation}
\argmin_R \| W_{t_1}R - W_{t_2} \|^2_F
\end{equation}

There is also an orthogonality constraint on the optimization of R such that R should satisfy $R^TR=I$. R can be obtained by first taking the SVD factorization of $M={W_{t_2}}^TW_{t_1}$ and then multiplying the left singular vectors $U$ and right singular vectors $V$, $R=UV^T$. This methodology makes diachronic analysis available on embeddings. Therefore, we provide aligned embeddings of consecutive time periods and a transformation matrix for each pair of time periods, along with synchronic embeddings for further analysis.

\section{Data Analysis} \label{sec_data_analysis}

In this section, we provide an analysis of our corpus to get more insight about it and how it can be utilized in future studies. In particular, we first provide statistical features of the corpus (Section \ref{sec_cor_stat}). Next, we diachronically analyze the corpus to understand how the Turkish language   has changed over time (Section \ref{sec_vocab_change}).

\subsection{General Statistics} \label{sec_cor_stat}

 \textbf{Table \ref{tab:descriptive_table}} provides the general statistics about \corpus{}. The dataset contains 
 45,375 documents and its size is 
 8.5 GB in text format. Before our filtering step, the dataset consists of around 
 849 million tokens 
 and the number of unique stems is 1,961,044.  
After the filtering process, the total number of tokens and unique stems were reduced to 842,957,298, and  211,775, respectively.

. 

  \begin{table}[!htb]
    \begin{tabular}{|p{0.6\linewidth}|r|}
    \hline
         The number of documents & 45375\\
         \hline
         The number of words before filtering & 849,335,014 \\
         \hline
         The number of words after filtering & 842,957,298\\
         \hline
         The number of unique surface level words & 10,689,405\\
         \hline
         The number of unique stems & 1,961,044\\
         \hline
        The number of unique stems after filtering & 211,775\\
         \hline
         
         Average token count per document & 18,718 \\
         \hline
    \end{tabular}
    \caption{Descriptive statistics of \corpus.}
    \label{tab:descriptive_table}
\end{table}

\noindent

\subsection{Vocabulary Change Across Years}\label{sec_vocab_change} 
 We first divide the documents into 10-year time period and compare each time period from different aspects. In all our calculations, we use lemmas (and stems for the words we the F5 stemming method). 
 
\begin{figure}[!htb]
    \centering
    \includegraphics[width=1\linewidth]{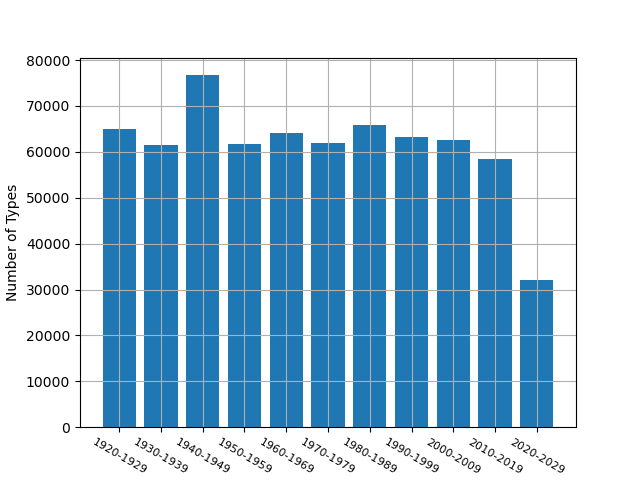}
    \caption{The number of unique lemmas/stems for each 10-years time period.}
    \label{fig:dia-surface_level_types}
\end{figure}

 Firstly, we compare vocabulary size across time periods. \textbf{Figure \ref{fig:dia-surface_level_types}} shows the number of words for each ten-year time period.  We observe that the vocabulary size is balanced in almost all time periods. Interestingly, the vocabulary size in 1940-1949 is higher than the others.  The vocabulary size for 2020-2022 is less than others due to the limited number of documents for this time period.
 
Next, we turn our attention to the vocabulary distance across different time-periods. In particular, we first create a separate list of unique words for each 10-year period. Next, we compute both Jaccard similarity and Jensen-Shannon Divergence (JSD) between documents in different time intervals. We chose Jaccard due to its high interpretability and JSD to better show how the vocabulary changes over time.


\begin{figure}[!tb]
    \centering
    \includegraphics[width=0.45\linewidth]
    {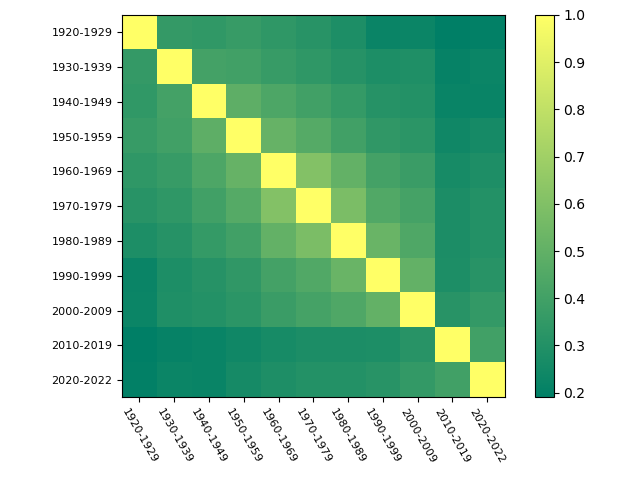}
    \includegraphics[width=0.45\linewidth]{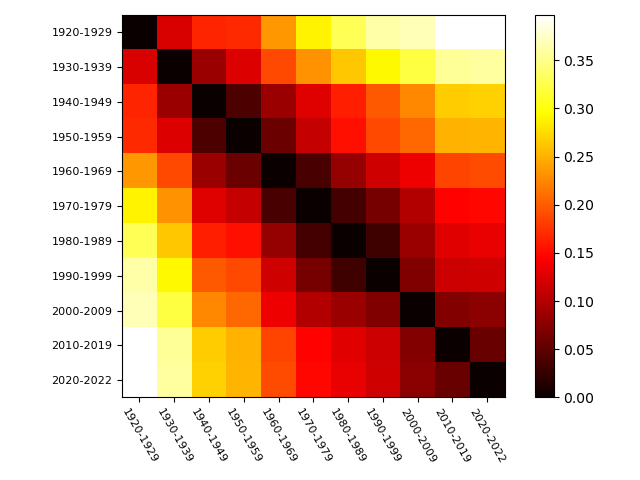}
    \caption{Jaccard similarity values of the vocabularies of 10-year periods.}
    \label{fig:jaccard-sim}
\end{figure}

\textbf{Figure \ref{fig:jaccard-sim}} shows the Jaccard similarity scores for the vocabulary of each 10-year period in heatmap format. We observe that the Jaccard similarity between two consecutive time periods is less than 50\% in  around half of the cases. Furthermore, the similarity between documents in the 1990s and documents in the 1920s is 
approximately 0.2\%. To illustrate this huge vocabulary change, we rewrite the first  sentence shown in Table \ref{tab_sample} using modern  Turkish words\footnote{the original words are written in parentheses, while their modern equivalents are highlighted in red}: \\

 \textit{\textcolor{red}{Egemenlik} (Hâkimiyet) \textcolor{red}{kayıtsız şartsız} (bilâ-kayd ü  
şart) milletindir.  \textcolor{red}{Yönetim} (idare) \textcolor{red}{şekli} (usulü), 
halkın \textcolor{red}{yazgısını} (mukadderatını) \textcolor{red}{doğrudan doğruya} 
(bizzat) ve \textcolor{red}{fiilen/gerçekten} (bilfiil) \textcolor{red}{yönetmesi} 
(idare etmesi) esasına \textcolor{red}{dayanmaktadır} (müstenittir)}.   
\textbf{Figure \ref{fig:heatmap}} shows the JSD scores between every pair of 10-year time periods. The first salient aspect of the heatmap is that as the distance between the compared time periods increases, the divergency increases in parallel. 

\begin{figure}[!htbp]
\hspace*{-2cm} 
    \includegraphics[width=1.27\linewidth, height=0.62\paperheight]{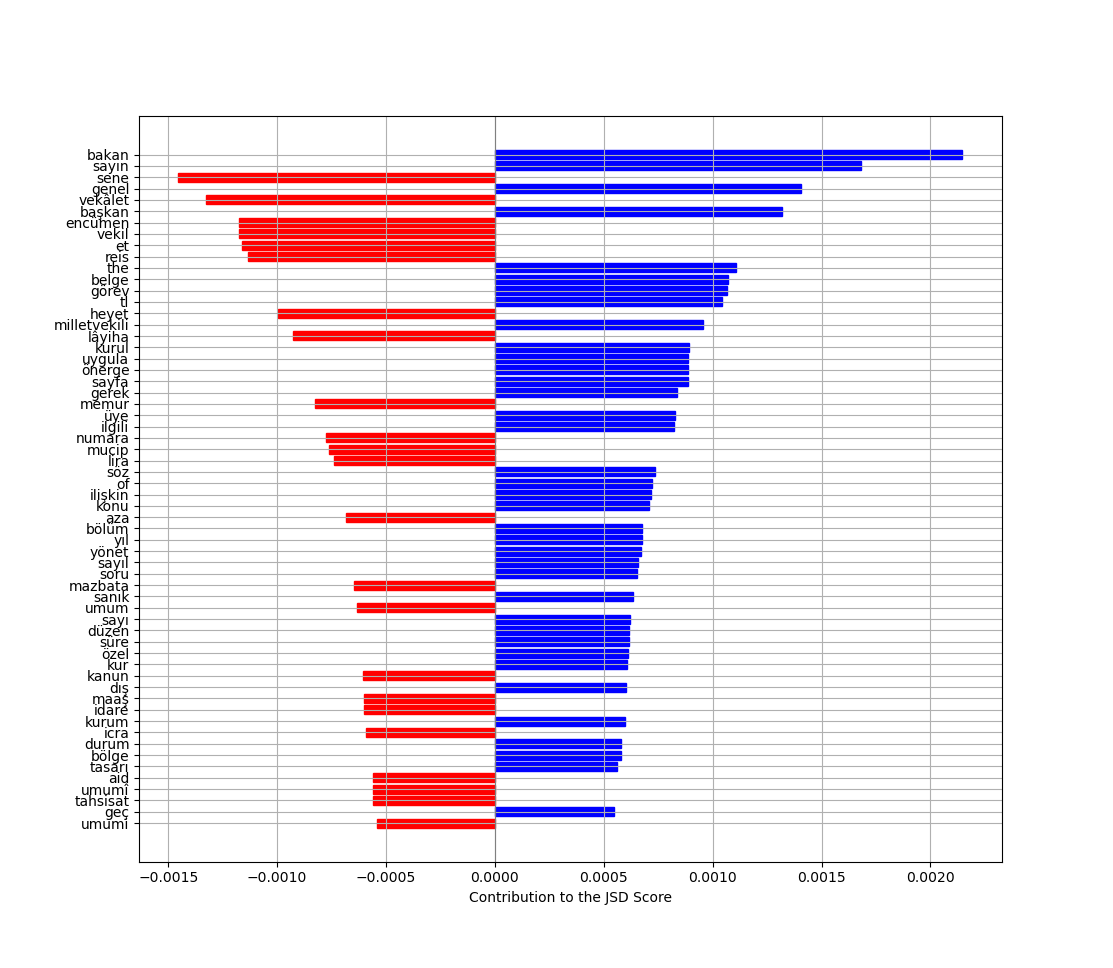}
    \caption{The first 60 words from 1930-1939 and 1980-1989 ordered by the individual contributions to the Jensen-Shannon divergence. The sign of the values indicates the corpus in which individual words are relatively frequent; the bars to the left represent the words that are more common in 1930-1939, while the bars to the right correspond to the words that are more frequent in 1980-1989.}
    \label{fig:divergent-words}
\end{figure}

In order to further investigate the vocabulary change, we rank the words based on their contribution to the JSD score. 
\textbf{Figure \ref{fig:divergent-words}} shows  the 60 words that cause the most divergency
between documents written in 1930-1939 and 1980-1989. We are not able to show the other comparisons due to the space limitation. 
In the figure,  the red and blue bars represent  words that are more characteristic in 1930-1939 and 1980-1989, respectively. The length of the bar indicates the magnitude of the contribution of each word to the overall JSD score.

\begin{table}[!htb]
\begin{tabular}{|l | l | l| l|}
\hline
\textbf{1930-1939} & \textbf{1980-1989}  & \textbf{Meaning}\\
\hline
vekil & bakan &  minister \\ \hline
sene & yıl &  year \\ \hline
umumi & genel &  general \\ \hline
reis & başkan &  president \\ \hline
heyet, encümen  & kurul &  committee \\ \hline
vesika & belge &  document \\ \hline
icra & uygula &  perform \\ \hline
mucip, lazım & gerek &  required \\ \hline
aza & üye &  member \\ \hline
idare (et) & yönet &  manage  \\ \hline
sayı & numara &  number \\ \hline
layiha & tasarı &  pleading \\ \hline

\end{tabular}
\caption{The words that have similar or identical meaning but were prevalent in different time periods.}
\label{tab_vocab}
\end{table}

Our results imply dramatic changes in Turkish vocabulary  in the last 100 years. 
 


\begin{figure*}[!htb]
    \centering
    \includegraphics[width=0.45\linewidth]{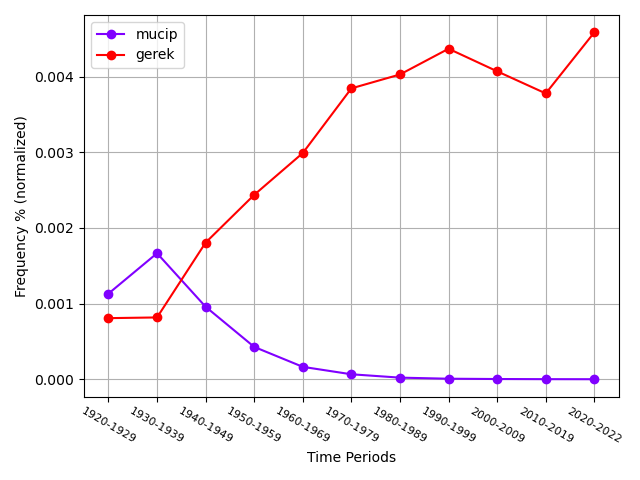}
    \includegraphics[width=0.45\linewidth]{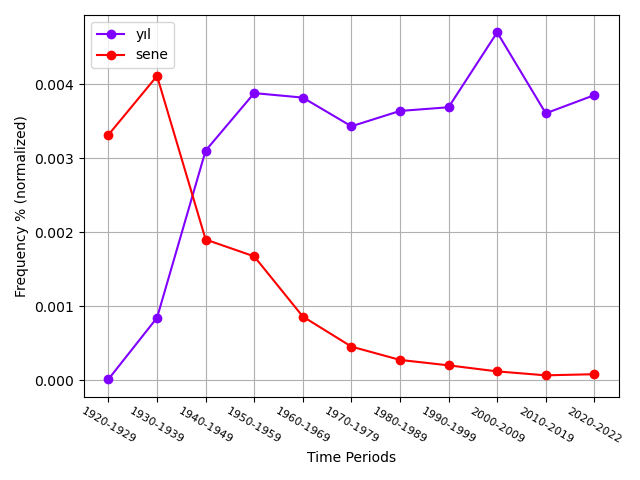}
    \caption{Normalized frequency of words that have the same meaning: gerek vs. mucip (required) and yıl vs. sene (year).}
    \label{fig:word}
\end{figure*}

In  Figure \ref{fig:divergent-words}, 
we observe that newly coined Turkish terms replaced the corresponding Arabic and Persian-origin words in the period of 1980-1989. \textbf{Table \ref{tab_vocab}} lists all the word pairs that have the same or similar meaning but appear in different time periods in Figure \ref{fig:divergent-words}.
For instance, both the words ``\textit{reis}'' and ``\textit{başkan}'' mean ``president'' in Turkish, but the word \textit{reis} is one of the most divergent words of the 1930-1939 period while  the word \textit{başkan} is one of the most divergent words of the 1980-1989 period. Furthermore, the words \textit{gerek} and \textit{lazım}  were used as replacements for the word \textit{mucip}  in 1980-1989 period. Moreover, the word \textit{kurul} was used to replace two different words, \textit{heyet} and \textit{encümen}.

We also observe that some words appear as divergent due to content or style change in the documents being compared.
For example, English words such as \textit{the} and \textit{of} are in the list of divergent words because international agreements and contracts are included in the documents of the 1980-1989 period. Regarding the style change, the words 
\textit{TL} which is the abbreviated version of \textit{Türk Lirası} (Turkish Lira) 
and \textit{lira} are two distinct terms from different time periods, indicating a shift towards adopting the abbreviation \textit{TL} for \textit{lira}. 

To shed more light on how words are introduced to replace Arabic-Persion origin words, we focus on two specific pairs of words with identical meanings: i) \textit{gerek} vs. \textit{mucip} (required) and ii)  \textit{yıl} vs. \textit{sene} (year). 
We calculate the frequency of each word and then normalize their frequency by the total number of tokens for each time period. \textbf{Figure \ref{fig:word}} shows the normalized frequency of these words. While both the words \textit{gerek} and \textit{mucip} (both means "required") existed in the 1920s, \textit{mucip}, which is an Arabic-origin word, is used more frequently than the word \textit{gerek}. However, in the following time periods the word \textit{gerek} becomes more popular than \textit{mucip}, and the word \textit{mucip} does not appear in the documents since the 1980s. 

In our second example, we observe another interesting case. The word \textit{yıl} does not exist in the  1920s in our corpus. However, it becomes popular to the extent that it is  more prevalent than the word \textit{sene} in all documents written after 1930s.  

\begin{figure}[!htb]
    \centering
    \includegraphics[width=\linewidth]{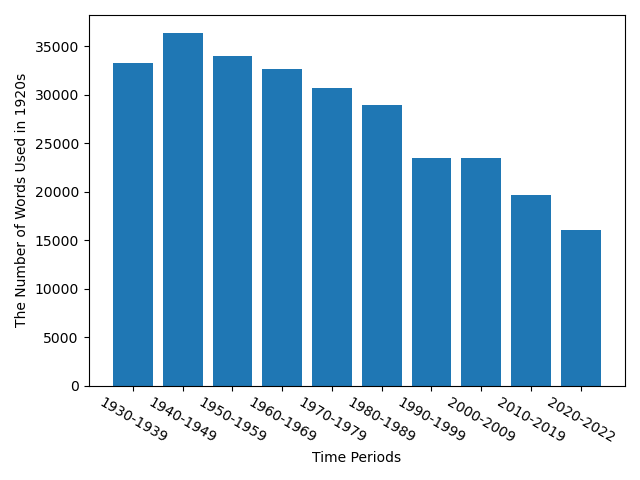}
    \caption{The number of words used in 1920s for each time-period.}
    \label{fig:survive}
\end{figure}

In our last analysis of vocabulary change, we investigate the presence of words used in the 1920s across subsequent time periods. \textbf{Figure \ref{fig:survive}} shows the number of words used in 1920s  for the subsequent time periods, i.e., \textit{survived} words. As expected, the number of survived words decreases as the time difference increases. Considering that there are around 
65,000 words used in documents of 1920s (See Figure \ref{fig:dia-surface_level_types}), around half of the words are not used in the subsequent years. 

\subsection{\textcolor{black}{Vocabulary Change on Diachronic Embeddings}}
\begin{table}[!htbp]

\begin{tabular}{|l | l |}
\hline
\textbf{1980-1989} & \textbf{1930-1939}\\
\hline
bakan & vekâlet, \textcolor{blue}{iktıs}, \textcolor{blue}{îktıs}, vekâl, içtimaî, maarif, \textcolor{red}{tktıs}, iktisat, nafıa, îcra\\
yıl & \underline{sene}, \underline{yıl}, yılma, ayı, takvim, aylık, seneye, iptida, \emph{dörd}, katıl \\
belge & \underline{vesika}, ibraz, istek, \textcolor{blue}{vesai}, makbu, musaddak, \textcolor{blue}{mütea}, \underline{vesaik}, talih, makbuz \\
gerek & icabe, \underline{iktiza}, göre, icap, ayrıç, kanunî, \underline{zarurî}, kabîl, \underline{lüzum}, önce \\
başkan & \underline{reis}, müteşekkil, seçim, \textcolor{red}{zatte}, müsteşar, inha, vekâlet, \textcolor{red}{mütal}, riyaset, seç \\
genel & müdür, \underline{umum}, işle, \underline{umumî}, idare, \textcolor{blue}{teşki}, denizyolu, genel, havayolu, îdare \\
kurul & seçim, \underline{heyet}, îcra, ödev, baro, seçilir, yönetim, inha, teşekkül, seç \\
uygula & \underline{tatbik}, göre, gözet, \textcolor{red}{dışmd}, hüküm, önce, \underline{tatbikat}, \textcolor{red}{icabl}, istisnaî, \textcolor{blue}{tatbi} \\
üye & seç, seçilir, seçilmiş, seçim, \underline{üye}, intihap, \textcolor{red}{seçi}, \textcolor{red}{zatte}, müntahap, müntehap \\ 
yönet & talimatname, talimat, nizamname, teşkilat, bölüm, \textcolor{blue}{teşki}, izahname, sayıl, ilgili, \textcolor{blue}{plânl} \\ 
numara & \underline{numara}, yazı, değiştiri, aşağı, ilişik, gösteri, \underline{sayı}, mezkûr, ün, yaz \\ 
tasarı & \textcolor{blue}{lâyih}, \underline{lâyiha}, \textcolor{red}{eneüm}, \textcolor{blue}{encüm}, değişik, mütenazır, tadil, bazı, encümen, ncü \\
\hline
\end{tabular}
\caption{The words that have similar or identical meaning but were prevalent in different time periods.}
\label{tab_vocab_emb}
\end{table}

We perform a diachronic analysis on word embeddings to further emphasize the change in the vocabulary, particularly the effect of the purification of Turkish. To be consistent with JSD analysis, we chose the same time intervals 1930-1939 and 1980-1989 using the same words listed in the second column of Table \ref{tab_vocab}. First, we train a CBOW model for both of the time periods. The configuration of the parameters of the models is specified in \ref{sec_embeddings}. After the models are trained, the word embeddings of 1980-1989 are aligned to 1930–1939 using the solution of the Orthogonal Procrustes method. Similarity scores between any two word vectors are calculated using cosine distance.  

Table \ref{tab_vocab_emb} shows some of the most characteristic divergent words of 1980-1989 and their 10 most similar words in 1930–1939. It can be observed that the old Arabic equivalents of the modern Turkish words are included in the closest 10 words as a result of the alignment of the word vectors. Underlined words are the near-synonyms of the words in the second column. Words colored with red stand for OCR errors, and words colored with blue are words that the morphological analyzer fails to stem; hence, they are the result of the F5 stemming process. We made this distinction since no word in the dictionary starts with words with red color, while words with blue color can be found as a prefix of a word in the vocabulary. Therefore, words blue words can be estimated from their similar words and the context. For example, in the first row of the table, the most similar 10 words of the old Arabic version of the modern Turkish word \textit{bakan} (minister) is in the last column. \textit{iktıs} is colored blue because it is probably the stemmed form of \textit{iktısad-} (economy). Note also that \textit{iktisat} which is an Arabic-originated word, is presented among the most similar words of \textit{bakan}. \textit{iktisat} and \emph{iktısad} corresponds the same meaning, \textit{ECONOMY}, and \textit{iktisat} is another form of \textit{iktısad}. Also, there is a third option for the meaning \textit{ECONOMY}: ekonomi. The word \textit{ekonomi} which originated from French, économie, was later introduced to replace  \emph{iktısad} \citep{sozluk} and prevails over the other two words, e.g in 1980-1989, the relative frequencies of \textit{iktisat} and \textit{ekonomi} are respectively $1.33\times10^{-5}$ and $16.9\times10^{-5}$, and \emph{iktısad} has become extinct. Furthermore, the surrounding words of \emph{iktisat} in 1930-1939, \textit{âli, celal, program (program), abdülhalik, vekâlet (ministry), nafıa (development), vekil (minister), millî (national), sıhhat (wellness), mustafa}, which is extracted with the help of PPMI matrix of 1930-1939, belong to the governmental context and \textit{ali, celal, abdülhalik, mustafa} are the first names of the authorities of the economy of Turkey. Whereas surrounding words of \emph{iktisat} in 1980-1989 are \emph{teori (theory), matematik (math), siyasal (political), kongre (congress), maliye (finance), politika (policy), teşebbüs (attempt), kıymet (value), teşekkül (organization)}. It can be observed that \emph{iktisat} has become used in the academic context as well.  Another example of the lexical change can be found by aligning the word vector of \textit{adet} (unit), which is not presented in the table \ref{tab_vocab}, to the vector space of 1930-1939. The most similar word from 1930-1939 is \textit{aded}. These results imply that such types of lexical change can be effectively discovered by examining diachronic word embeddings.

Diachronic embeddings have another interesting property. Although, a word or a concept is not present in a vocabulary, aligning a word from a different time period may reveal related words. To be more specific, words similar to the aligned word are often related to the relevant concept. As an example from our corpus, the word \textit{televizyon} (television) is not present in the vocabulary of 1930-1939, and TELEVISION was not a well-known concept in these years. We aligned the vector of \emph{televizyon} from 1980-1989 to the vector space of 1930-1939. The most similar 10 words of the aligned vector of \emph{televizyon} are radyo (radio), sinema (cinema), tiyatro (theater), mecmua (magazine), arsıulusal (international), rehber (guide), broşür (brochure), reklâm (advertisement), telsiz (radio), konser (concert). Most of the similar words belong to the concept MEDIA as does \emph{televizyon}.

\subsection{Change in Writing Conventions} 

We conduct two different analyses for changes in writing conventions. We first explore how the word endings have changed and then we focus on the usage of circumflexes.

\begin{figure}[!htb]
    \centering
    \includegraphics[width=\linewidth]{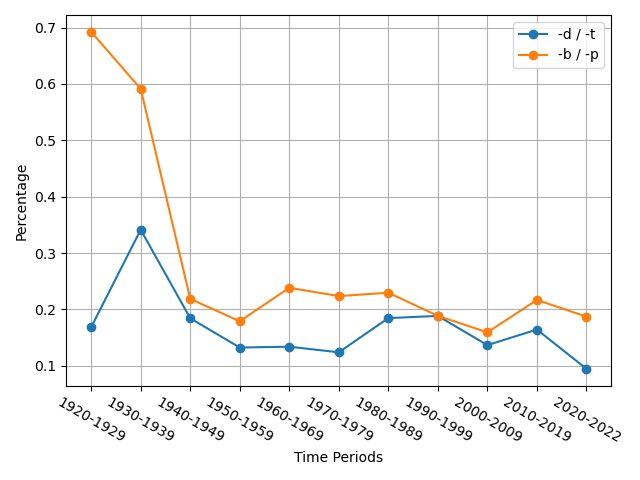}
    \caption{The ratio of words ending with -b/-d letters compared to the words ending with -p/-t letters in each time period.}
    \label{fig:endling_letter}
\end{figure}

The words originating from Turkish do not end with the letters "-b", "-c", "-d", and "-g" with few exceptions. However, there are several loanwords from Arabic and Persian that end with one of these letters. We observe that these names have been written in two different ways in which the last letter is changed from  "-b/c/d/g" to "p/ç/t/k" such as \textit{Ahmet} vs. \textit{Ahmed}. 
In order to observe the transformation of these loanwords, we first detect words in which a single letter is different based on this phonetical rule, e.g., Ahmet vs. Ahmed\footnote{We removed the word \textit{et} (which is an auxiliary verb) because of its extremely high prevalence compared to others.}.  Next, we count the words that end with -d and -b  and their counterparts ending with the letter -t and -p for each 10-year time period. We ignore words ending with -c/-ç and -g/-ğ letters due to their low prevalence. 
Subsequently, we calculate the ratio of words ending with -b/-d letters compared to the words ending with -p/-t letters.  The results are shown in \textbf{Figure \ref{fig:endling_letter}}.


We observe that the ratio for both letter pairs is less than 1, suggesting that ending words with "-t" and "-p" letters is more common than using "-d" and "-b" letters in all time periods. However, this might be because of the morphological analysis tool we use. In particular, as it is developed based on modern Turkish grammar rules, it might identify stems as if they end with "-t" and "-p" in some cases. Therefore, it is important to focus on the trend instead of actual values. 

The percentage of words ending with "-b" significantly decreases between 1920s and 1940s. Afterwards, 
it fluctuates between the levels of 0.16 and 0.24. However, we observe a different pattern for words ending with "-d". Interestingly, the percentage of words ending with "-d" letter first increases (from 1920s to 1930s) and then decreases (from 1930s to 1940s). In the following time periods, it fluctuates between 0.1 and 0.2. These fluctuations might stem from the limitations of the corpus or due to people's resistance to reforms in language. For instance, it is still common that many people in Türkiye give names ending with letter "-d"  to their children. 

\begin{figure}[!htb]
    \centering
    \includegraphics[width=\linewidth]{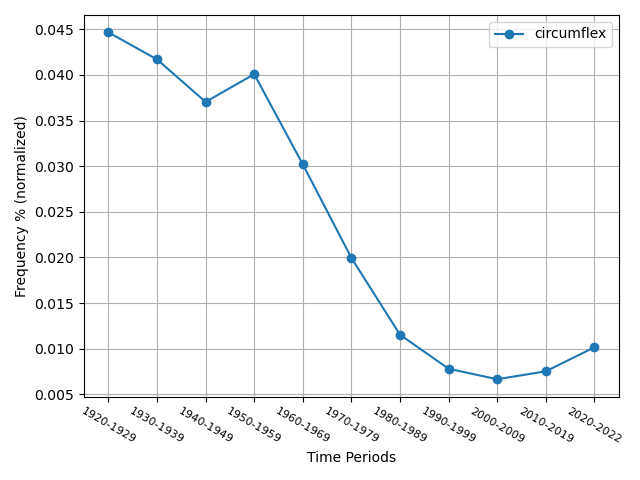}
    \caption{The frequency of the usage of the circumflex in our corpus.}
    \label{fig:caret_usage}
\end{figure}

An interesting issue with the spelling changes in the Turkish language is the urban legend about the removing of letters with a circumflex (\textasciicircum). In particular, some letters like "-a", "-{\i}", and "-u" are written with a circumflex in some words, e.g.,  \textit{kâğıt} (paper), \textit{abidevî} (monumental), \textit{şûra} (council). While these letters are not removed from the official alphabet, many people on social media platforms claim that it was first removed but brought back later on. Even fact-checking websites had to verify the veracity of the claim\footnote{https://www.malumatfurus.org/sapka-isaretinin-kaldirildigi-iddiasi/}. In our corpus, we also focus on this urban legend and count the number of letters with a circumflex. The results are shown in \textbf{Figure \ref{fig:caret_usage}}. We observe that letters with a circumflex have varying frequencies over time but they are continued to be used. However, we also notice that their frequencies have significantly dropped, which might be the reason to have such an urban legend.

\section{Limitations}\label{sec_limitations}

While our work provides valuable insights into how the Turkish language has changed since the 1920s, there are particular issues that need to be taken into account when analyzing our results. 
Our corpus mainly represents the language used by authorities on topics about governing. Therefore, it does not represent the whole characteristics of the Turkish language.  That said, one of the main reasons for the relatively rapid evolution of the Turkish language is the government's policies aimed at language reform such as proposing new words to replace Arabic or Persian origin words and changing the alphabet. Therefore, our corpus might be one of the best resources to investigate the intervention of government in the Turkish language reform.

In our study, we use tools to automatize text extraction from the PDF files and detect lemmas of words. The tools we use might introduce noise and affect our results. \textcolor{black}{Especially, in the diachronic analysis of embeddings, we aligned the consecutive time periods and measured the semantic change. We see that the most semantically displaced words are the noise words that survived throughout time periods. Also, in the word similarity task, we see that a set of k-similar words to a word contains OCR errors, and the errors follow some specific patterns, e.g. \emph{-ü} has been scanned as \emph{-ii} or \emph{-e} has been taken as \emph{-c}. Such errors mislead the embedding models in the training phase.}. Therefore, we take action to reduce noise and its impact. To further mitigate this problem, we also share our code and dataset, increasing the reproducibility of our findings and enabling further research on this dataset. Nevertheless, the possible impact of the tools we use should be taken into account when analyzing our findings.

\section{Conclusion}\label{sec_conclusion}

 The Turkish language has encountered multifaceted transformation over the last century, underscored by state-driven initiatives such as changing the alphabet and replacement of loanwords with Turkish-origin words. In order to enable future studies on this interesting linguistic transformation of Turkish, we introduce  \corpus{}; \textcolor{black}{a toolkit that comprises various types of Turkish diachronic resources such as raw text corpus extracted from the Official Gazette of Türkiye and the records of Grand National Assembly of Türkiye over a century-long, diachronically aligned embeddings of different kinds, collocation matrices, a digitized dictionary of modern-old correspondent Turkish words, and Python library that allows diachronic analysis.}  
 Next, we conduct a comprehensive diachronic analysis using our corpus to investigate language reform in Turkish.  In particular, we first explore how the vocabulary has changed since 1920. Next, we investigate how the spellings of words have changed. 

 Based on our comprehensive analysis, our findings are as follows. i) The vocabulary has dramatically changed throughout the years such that almost half of the words used in the 1920s were not used in the 2010s. We observe that the frequency of loanwords decreases while the frequency of words used for replacement increases throughout the years. Regarding the changes in spelling, our analysis reveals a noticeable decline in the use of circumflex compared to the 1920s and 1930s. Furthermore, there has been a shift in the final letters of words over time, influenced by Turkish phonology. Specifically, words ending in "-b"  have decreased over time in favor of versions ending in "-p". However, a distinct pattern emerges for words ending with "-d" or "-t" letters: The proportion of words ending with the letter "-d" compared to those ending with the letter "-t" remains similar to the proportions seen in the 1920s, although there has been a declining trend since the 1990s.

 We think that \corpus{} paves the way for targeted studies on specific linguistic phenomena in Turkish, such as the evolution of certain grammatical structures, lexical borrowing, or semantic drift over time. While the introduction of a diachronic corpus for Turkish fills a significant gap in linguistic research,  we plan to extend our corpus by the inclusion of texts from other sources, such as newspapers, literary works, and public broadcasts from corresponding periods in the future. Once we build a larger diachronic corpus from various sources, we plan to extend our analysis on Turkish language reform and compare the differences across different data sources. 
 In addition, we plan to develop a software enabling easy access and analyze of the corpus for researchers. 

\bibliography{sn-bibliography}

\end{document}